\documentclass[review]{elsarticle}

\usepackage{hyperref}
\usepackage{graphicx, float}
\usepackage[cmex10]{amsmath}
\usepackage[numbers]{natbib}
\usepackage[usenames, dvipsnames]{color}

\journal{Engineering Applications of Artificial Intelligence}

\begin{document}

\begin{frontmatter}

\title{Optimizing Speed/Accuracy Trade-Off for Person Re-identification via Knowledge Distillation}

\author[UTRCadress]{Idoia Ruiz\corref{mycorrespondingauthor} \fnref{myfootnote}}
\fntext[myfootnote]{Present address: Computer Vision Center, Universitat Aut\`{o}noma de
Barcelona, Edifici O, 08193 Bellaterra, Spain.}
\cortext[mycorrespondingauthor]{Corresponding author}
\ead{iruiz@cvc.uab.es}

\author[CVCadress]{Bogdan Raducanu}
\ead{bogdan@cvc.uab.es}

\author[UTRCadress]{Rakesh Mehta}
\ead{mehtar1@utrc.utc.com}

\author[UTRCadress]{Jaume Amores}
\ead{amoresj@utrc.utc.com}

\address[UTRCadress]{United Technology Research Centre Ireland, 4th Floor, Penrose Business Center, Penrose Wharf, Cork City, Co. Cork, Republic of Ireland}
\address[CVCadress]{Computer Vision Center, Universitat Aut\`{o}noma de Barcelona, Edifici O, 08193 Bellaterra, Spain.}

\begin{abstract}
Finding a person across a camera network plays an important role in video surveillance. For a real-world person re-identification application, in order to guarantee an optimal time response, it is crucial to find the balance between accuracy and speed. We analyse this trade-off, comparing a classical method, that comprises hand-crafted feature description and metric learning, in particular, LOMO and XQDA, to deep learning based techniques, using image classification networks, ResNet and MobileNets. Additionally, we propose and analyse network distillation as a learning strategy to reduce the computational cost of the deep learning approach at test time. We evaluate both methods on the Market-1501 and DukeMTMC-reID large-scale datasets, showing that distillation helps reducing the computational cost at inference time while even increasing the accuracy performance.
\end{abstract}


\end{frontmatter}

\section{Introduction}
Person re-identification refers to the problem of identifying a person of interest across a camera network \cite{PRW,Panda_2017_CVPR}. This task is specially important in surveillance applications, since nowadays the security systems in public areas such as airports, train stations or crowded city areas, are continuously improving to ensure the population's welfare. In big cities, there are extensive networks of cameras in the most sensitive locations. 
Identifying an individual requires finding it among all the instances that are present on the collection of images captured by the cameras. These images show usually complex crowded scenes, thus increasing even more the computational complexity of the problem.
Therefore, the automation of this task that involves large-scale data becomes essential, as otherwise it would be a laborious task to be performed by humans.

\begin{figure}[H]
\begin{center}
 \includegraphics[scale = 0.35]{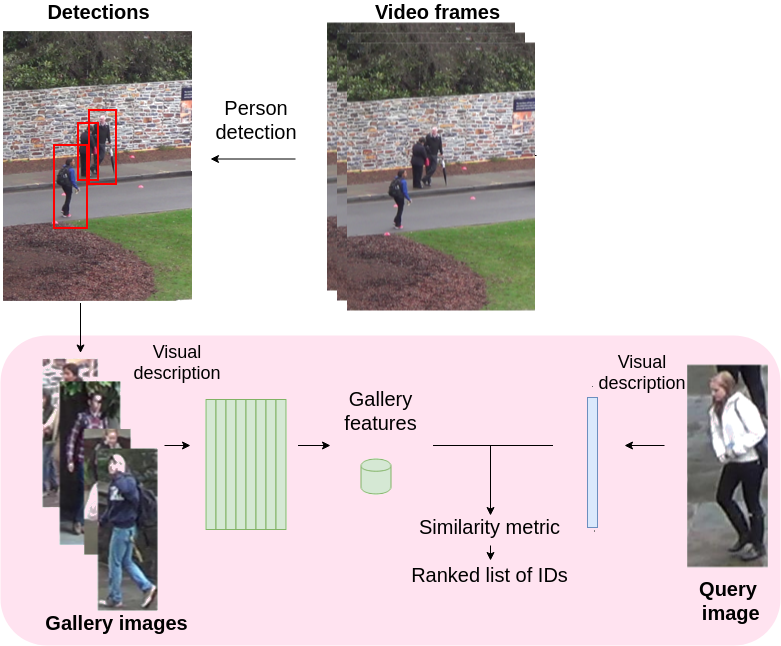}
\end{center}
\caption{Pipeline of an end-to-end person re-identification system. The pink shaded region delimits the person re-identification module.}
\label{fig:pipeline}
\end{figure}

The aim of person re-identification is to find a person of interest, also referred as \textit{query}, across a \textit{gallery} of images.
The difficulty of this problem lies in the fact that the images are subject to variations in the point of view, person pose, light conditions and occlusions. Fig. \ref{fig:datasets} shows examples of gallery images for identities with such kind of variability.
Fig. \ref{fig:pipeline} shows the full person re-identification system, including the previous person detection stage. 
In the person re-identification module, a \textit{query} image of a person of interest is compared against the \textit{gallery}, retrieving the images that correspond to the same identity. 
To compare them, the system first extracts a feature representation that describes every image, either by using a hand-crafted descriptor or a deep neural network. Usually the features of the \textit{gallery} are previously computed offline and stored, so that at test time we only have to extract the features for the query image. Once the features are extracted, they can be compared with the features of the \textit{gallery} by computing a similarity measure. Finally, all the gallery images are ordered by the degree of similarity, obtaining a ranked list of the most similar images in the gallery to the person of interest \cite{Zhong_2017_CVPR}.

In real scenarios, in order to have a feasible application that is able to work with large-scale datasets in an efficient and effective way, we have to address the problem of optimizing the computational cost of the system at test time, without decreasing drastically its accuracy. For that purpose, we consider both classical and deep learning based person re-identification methods. Although deep learning based techniques outperform significantly hand-crafted methods in terms of accuracy, their drawback is that they require dedicated hardware, \textit{i.e.} GPUs, and big amounts of data for training, which takes usually long periods of time, \textit{i.e.} weeks, in order to be effective. 

To make deep learning approaches computationally efficient several works use model compression \cite{buciluǎ2006model,ba2014deepnets}. The idea behind model compression is to discard \emph{non-informative} weights in the deep networks and perform a fine-tuning to further improve performance. Although these methods make the architecture more efficient in terms of computational complexity,  they also result in a drop of the accuracy on the compressed models. This drop is specially prominent when the dataset is large or the number of classes is higher, which is often the case in the person re-identification problem. In contrast, network distillation works have shown that the smaller or compressed model trained with the support of a much bigger/deeper network is able to achieve very similar accuracy as the deeper network but having a much lower complexity \cite{romero2014fitnets,zagoruyko2017attentionmaps,ba2014deepnets}. Therefore, in this work we explore network distillation in the context  of efficient person-re-identification. 

\paragraph{\textbf{Contribution}} The goal of this work is first, to provide an analysis of the trade-off between accuracy and computational cost at test time in a person re-identification problem, considering the most suitable configuration for a real-world application conditions, and second, to propose an improvement to optimize this trade-off.
The contribution of this work is, first, to provide such trade-off analysis on two challenging large-scale person re-identification benchmarks, that are Market-1501 \cite{market1501} and DukeMTMC-reID \cite{Duke}, and finally to introduce and analyse network distillation \cite{hinton2015distilling} for optimizing this trade-off for the deep learning approach. 
For this purpose, we use ResNet-50 \cite{Resnet}, acting as teacher, to transfer the knowledge to a more compact model represented by MobileNet \cite{mobilenets}, acting as student.

The paper is structured as follows. In Section 2, we review the literature related with person re-identification and distillation. In Section 3, we review the distillation approach. The experimental results are reported in Section 4. Finally, in Section 5, we present our conclusions and provide some guidelines for future work.

\section{Related Work}
\label{sec:SOA}

\subsection{Person re-identification}
\label{sec:SOA_prid}

Classical methods for person re-identification consider it as two independent problems, that are feature representation and metric learning.
For the first task, visual description, popular frameworks like Bag of Words \cite{zheng2009associating} or Fisher Vectors \cite{perronnin2010large} were initially used to encode the local features.
Later, the LOMO \cite{LOMO} descriptor was introduced and commonly used on the person re-identification problem \cite{varior2016siamese} \cite{Zhong_2017_CVPR} \cite{Panda_2017_CVPR}.
In the exhaustive comparison performed by \citet{karanam2016systematic}, LOMO is the second hand-crafted feature descriptor that performs best across several datasets. The GOG \cite{GOG} features are superior in terms of accuracy, but computing them is more computationally expensive, since it requires modeling each subregion in which the image is divided, by a set of Gaussian distributions. Indeed, in \cite{GOG}, LOMO features are extracted in 0.016 seconds/image, while GOG features are extracted in 1.34 second/image.

Metric learning consists in learning a distance function that maps from the feature space to a new space in which the vectors of features that correspond to the same identity are close, while those that correspond to different identities are not, being the distance a measure of the similarity.  Once learnt, this mapping function is used to measure the similarity between features of the person of interest and the gallery images.

One of the most popular metrics is KISSME \cite{KISSME}, that uses the Mahalanobis distance.
Later, XQDA\cite{LOMO} was introduced as an extension of KISSME to cross-view metric learning, but doing the mapping function from the feature space to a lower dimensionality space, in which the similarity metric is computed.
More recently, \cite{ali2018eccv} proposed a novel metric learning method that address the small sample size problem, which is due to the high dimensionality of the features on person re-identification. According to this metric, the samples of distinct classes are separated with maximum margin while keeping the samples of same class collapsed to a single point, to maximize the separability in terms of Fisher criterion.

Nowadays, deep learning based methods are outperforming hand-crafted techniques. Some approaches used deep learning to compute better image representations, then computing the similarity metric as usual.
Considering each identity as a different class, the features are extracted from a classification Convolutional Neural Network (CNN), that is trained on the target dataset. Then the features, that we denote as \textit{deep features}, are the logits, \textit{i.e.} the output of the network before the classification layer.
Some works that use this approach are \cite{PastPresentFuture}, \cite{xiao2016learning} and \cite{Bak_2017_CVPR}.
A more complex framework is proposed in \cite{Li_2017_CVPR}, where using a multi-scale context-aware network, they compute features that contain both global and local part-based information.

In a different line of work, siamese models were used to learn jointly the representations, computing the similarity between the inputs, that are image pairs. The similarity measure provided by the output of the network, determines whether the input images correspond to the same identity or not.
This architecture was first introduced by \citet{siamese} for signature verification, where the features for two signature images were extracted and compared by computing the cosine of the angle between the two feature vectors as a measure of the similarity. Similarly, in person re-identification, siamese networks take as an input two person images. This original approach is followed in \cite{yi2014deep}. Other architectures such as \cite{cuhk03} or \cite{ahmed2015improved} use the softmax layer to provide a binary output.
A siamese framework is also used in \cite{radke2019cvpr}, where the authors propose an architecture with an enhanced attention mechanism, in order to increase the robustness for cross-view matching.
Closely related to siamese networks, triplet networks, which were introduced in \cite{schroff2015facenet} for face recognition, take triplets of images as inputs, corresponding only two of them to the same person \cite{cheng2016person,Zhang2019LearningIT,Zheng_2019_CVPR}. Similarly, a quadruplet loss was proposed in \cite{Chen_2017_CVPR}. 

Recent approaches aim at increasing the robustness of person re-identification systems.
Some address the problem of domain adaptation, \textit{i.e.} applying to an unseen dataset a model is trained on a set of source domains without any model updating  \cite{hospedales2019cvpr,Liu_2019_CVPR}. To this end, image synthesis \cite{jiao2018cvpr,zhong2018eccv} or domain alignment \cite{kot2018bmvc,li2018cvpr,tian2018cvpr} are used.
Other works propose generative approaches for data augmentation. 
In \cite{xue2018eccv} the synthesized images help learning view-point invariant features by normalizing across a set of generated enhanced pose variations, while in \cite{kautz2019cvpr} they compose high-quality cross-identities images.

\subsection{Network Distillation}
Network distillation approaches appeared as a computational effective solution to transfer the knowledge from a large, complex neural network (often called \textit{ teacher network}) to a more compact one (referred as \textit{student network}), with significantly less number of parameters.
This idea was originally proposed in \cite{hinton2015distilling}. On their approach, the student network was penalized based on a softened version of the teacher network's output. The student was trained to predict the output of the teacher, as well as the true classification labels.
In \cite{romero2014fitnets}, they proposed an idea to train a student network which is deeper and thinner than the teacher network. They do not only use the outputs, but also the intermediate representations learned by the teacher as hints to improve the training process and final performance of the student.
A different approach was proposed in \cite{luo2016neurons}, where the knowledge to be transferred from the teacher to the student is obtained from the neurons in the top hidden layer, which preserve as much information as the softened label probabilities, but being more compact.

Network distillation approaches have also been applied recently to the person re-identification problem.
In \cite{lu2018cvpr}, the authors propose using a pair of students to learn collaboratively and teach each other throughout the training process. Each student is trained with two losses: a conventional supervised learning loss, and a mimicry loss that aligns each student’s class posterior with the class probabilities of other students. This way, each student learns better in such peer-teaching scenario than when learning alone.
In \cite{ge2018nips}, feature distillation is used to learn identity-related and pose-unrelated representations. They adopt a siamese architecture, consisting each branch of an image encoder/decoder pair, for feature learning with multiple novel discriminators on human poses and identities.
The recent work in \cite{lai2019cvpr} resembles ours in some aspects, although their scope is semi-supervised and unsupervised person re-identification, in contrast to our fully-supervised formulation. Similarly to us, they consider lightweight models to reduce testing computation as well as network distillation as an strategy of knowledge transfer. However, their distillation approach is not probability based, but similarity based. They propose the Log-Euclidean Similarity Distillation Loss that imitates the pairwise similarity of the teacher instead of using soft labels as we do. They explore a multiple teacher-single student setting and propose an adaptive knowledge aggregator to weight the contributions of the teachers.

\section{Reviewing Distillation}
\label{sec:method}
Besides improving the performance of the person re-identification pipeline in terms of computational cost at test time, we also aim at maximising the performance of a small network to be as accurate as possible.

\begin{figure}[h]
  \centering
  \centerline{\includegraphics[scale = 0.3]{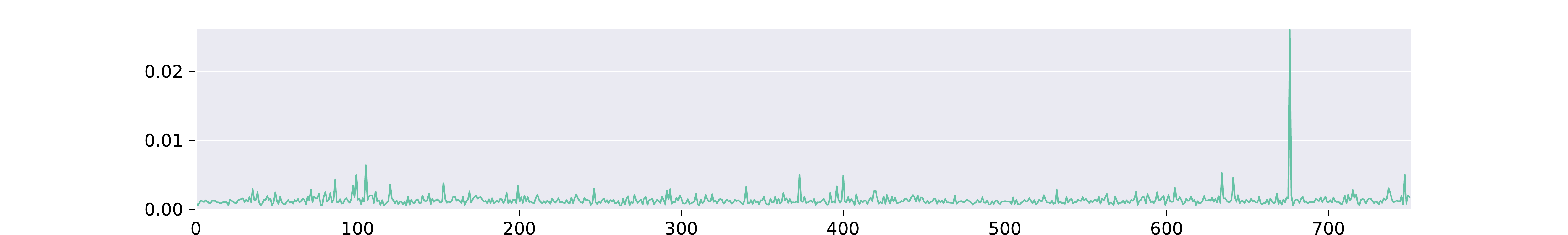}}
\caption{Example of a high entropy probability distribution, generated by the softmax layer of the teacher network for the Market-1501 dataset (751 classes). }
\label{fig:entropydist}
\end{figure}

As discussed in \cite{hinton2015distilling}, the simplest way to transfer the knowledge is to use the output of the teacher network as soft targets for the student network, additionally to the hard targets provided by the ground truth.
However, when the soft targets have high entropy, they provide more information to learn from. Then, a network that is very confident about its prediction, will generate a probability distribution similar to a Dirac delta function, in which the correct class has a very high probability and the rest of classes have almost zero probability, having a very low entropy and consequently providing less information than a less confident network. While a less confident network will assign higher probabilities to the incorrect classes, as shown graphically in Fig. \ref{fig:entropydist}. The intuition behind high entropy distributions help the distillation, is that by learning from the probabilities assigned to incorrect classes, the student network is learning how the teacher model generalizes.

Therefore, the authors propose to increase the entropy of the probability distribution generated by the teacher model, \textit{i.e.} the output of the softmax layer, so that when the student network uses that output to learn from it, it can provide more information. In order to maximize the entropy, they propose to increase the \textit{temperature} of this distribution.

The inputs of the softmax layer, that are the \textit{logits}, denoted as $z_i$, are converted to probabilities $p_i$ by the softmax function, which expression is (\ref{eq:softmax}), where T is the temperature, that is a selected constant value in the distillation case, and it is equal to 1 when there is no distillation.

\begin{equation}\label{eq:softmax}
p_i = \frac{exp(z_i/T)}{\sum_j exp(z_j/T)}
\end{equation}

The knowledge transfer is performed via the loss function of the student model. The loss function for the k-th training example $L_{student_k}$ is defined as (\ref{eq:lossdistill}) and it is the weighted sum of two terms:

\begin{equation}\label{eq:lossdistill}
\begin{split}
L_{student_k} \; = \underbrace{\; H\big(p_{teacher}(T=T_0), p_{student}(T=T_0)\big)}_\text{Distillation term} \; 
+\\ \; \lambda \underbrace{H\big(hard\; targets, p_{student}(T=1)\big)}_\text{Cross-entropy loss}
\end{split}
\end{equation}

where $H(p,q)$ denotes the cross-entropy between two probability distributions p and q. The first term is the cross-entropy between the soft targets extracted from the teacher ($p_{teacher}(T=T_0)$), \textit{i.e.} the softened probability distribution of the teacher that is obtained by applying the softmax function (\ref{eq:softmax}) to the logits of the teacher divided by a temperature $T_0$, and the softened probability distribution of the student ($p_{student}(T=T_0)$) using the same value $T_0$ as for the teacher. The second term of the loss is the cross-entropy between the \textit{hard targets}, that is the ground truth which has a value equal to 1 assigned to the correct class and 0 to the rest of them, and the probability distribution of the student ($p_{student}(T=1)$), that is the output of the softmax using a $T=1$. This second term is the cross-entropy loss function, which minimizes the cross-entropy between the prediction of the network and the ground truth. These two terms are balanced by a regularization parameter $\lambda$.

\begin{figure}[H]
  \centering
  \centerline{\includegraphics[scale = 0.45]{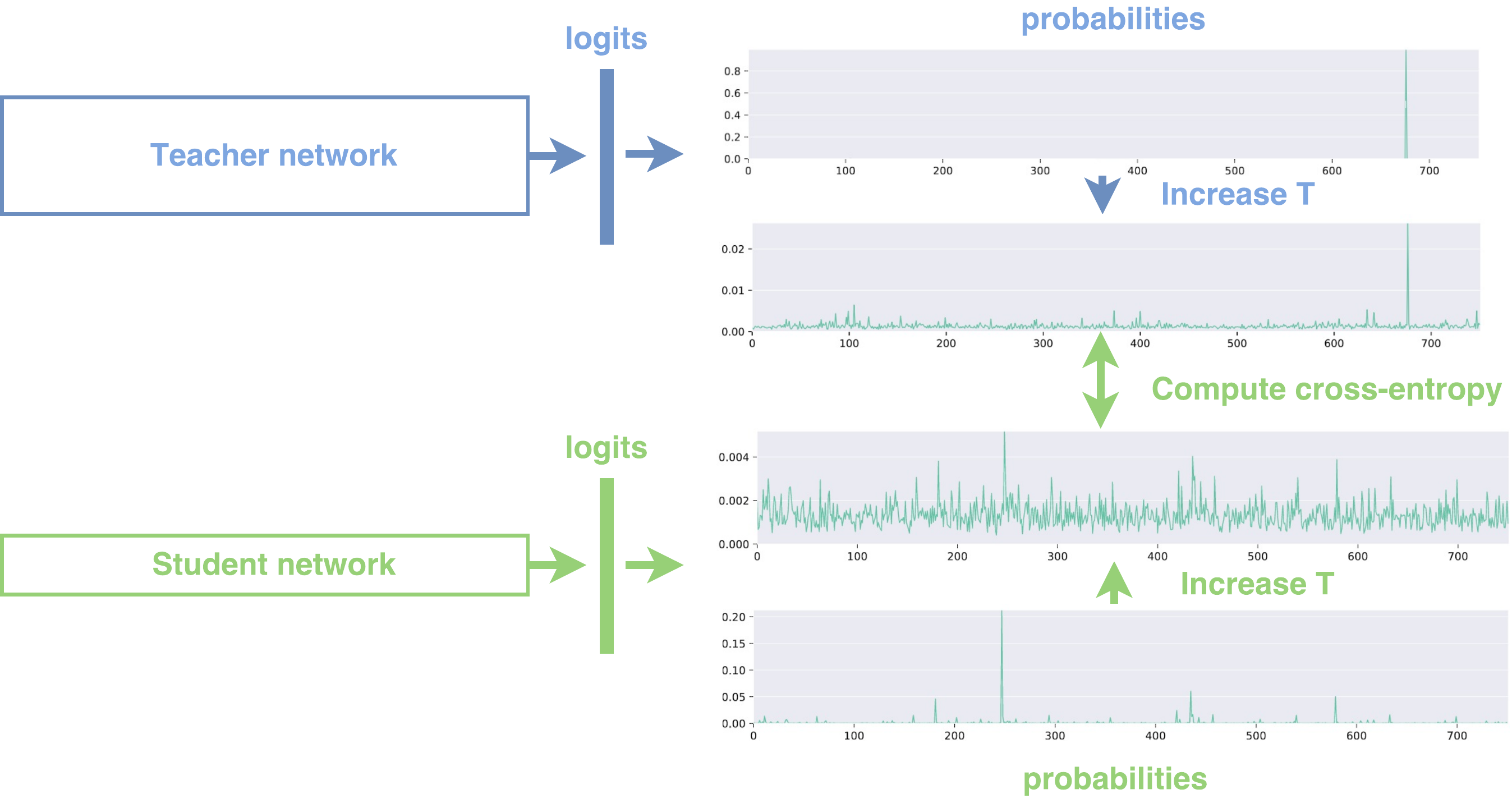}}
\caption{Distillation process. The cross-entropy between the softened distributions generated by the teacher and the student networks is computed in order to minimise it additionally to the cross-entropy with the ground truth.}
\label{fig:distillationprocess}
\end{figure}

A graphical summary of the process is shown in Fig. \ref{fig:distillationprocess}. 
In the current framework of person re-identification, once the student network is trained via distillation, it is used to extract the features of the images at test time, to then measure their similarity using the Euclidean distance.

\section{Experiments}
\label{sec:exp}
\subsection{Datasets}
\begin{figure}[H]
\begin{minipage}{.4\linewidth}
  \centering
  \centerline{\begin{tabular}{ccccc}
\includegraphics[width=0.13\textwidth]{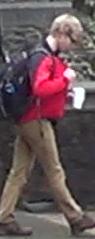} &
\includegraphics[width=0.13\textwidth]{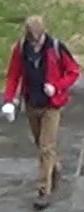} &
\includegraphics[width=0.13\textwidth]{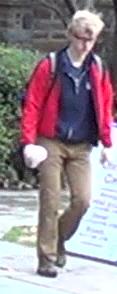} &
\includegraphics[width=0.13\textwidth]{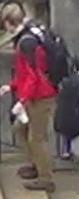} &
\includegraphics[width=0.13\textwidth]{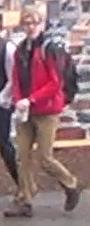}  
\end{tabular}}
  \centerline{(a)}
\end{minipage}
\hfill
\begin{minipage}{0.4\linewidth}
  \centering
  \centerline{\begin{tabular}{ccccc}
\includegraphics[width=0.15\textwidth]{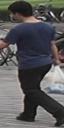} &
\includegraphics[width=0.15\textwidth]{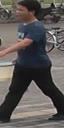} &
\includegraphics[width=0.15\textwidth]{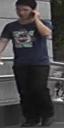} &
\includegraphics[width=0.15\textwidth]{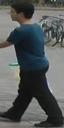} &
\includegraphics[width=0.15\textwidth]{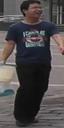} 
\end{tabular}}
  \centerline{(b)}
\end{minipage}
\caption{Subset of gallery images that correspond to 2 identities from the (a) DukeMTMC-reID and (b) Market-1501  datasets. Each identity can appear in different cameras and may present different points of view, pose, and illumination conditions.}
\label{fig:datasets}
\end{figure}

In a real world application, there are often several cameras that can capture images of people from different points of view in different illumination conditions and even with occlusions. 
Thus, we choose datasets that simulate as much as possible a real scenario.
Market-1501 \cite{market1501} or DukeMTMC-reID \cite{Duke} have these characteristics, providing images taken from 6 cameras in the case of Market-1501 and 8 in the case of DukeMTMC-reID, as shown in Fig. \ref{fig:datasets}, that are captured in outdoor public areas, being also two of the largest-scale public datasets for person re-identification. 

Market-1501 provides an average of 14.8 cross-camera ground truths for each query, containing in total 32,668 bounding boxes of 1,501 identities, from which 12,936 bounding boxes with 751 identities belong to the training set. The mean of images per identity is 17.2. All the bounding boxes are of size 128x64. 

The DukeMTMC-reID dataset is an extension of the DukeMTMC tracking dataset. The bounding boxes are then extracted from the full frames provided by the original dataset and therefore, their size is not fixed. It contains 36,441 bounding boxes that belong to 1,404 identities plus 408 distractor identities that only appear in a single camera. Among them, 16,522 bounding boxes with 702 identities are used for the training set. The mean number of images per identity is 20, with a maximum of 426 images for the identity with the largest amount of images.

\subsection{Evaluation}
In a re-identification task, the \textit{query} is compared to all the \textit{gallery}, computing a similarity metric that is used to rank the \textit{gallery} images sorted by similarity.
The rank-1 accuracy gives the probability of getting a true match from the gallery in the first position of the ranking. Similarly, the rank-5 accuracy evaluates whether we find a true match in the five first positions of the ranking.
However, since the person of interest may appear many times in the gallery, we need an evaluation metric that also considers finding all the true matches that exist in the gallery, evaluating also the recall. The mean average precision (mAP) is suitable for evaluating on datasets in which an identity appears more than once in the gallery, such as Market-1501 and DukeMTMC-reID.

We also report the computational cost at test time of the algorithms proposed, by providing the time that feature extraction takes per image of a single individual. We do the feature extraction for all the \textit{gallery} and compute the average time per image. We report as a computational cost metric, the number of images the system extracts the features from in a second, for the different considered architectures. Then, we report separately the computational cost for the metric learning step.

\subsection{Implementation details}
\label{sec:Exp_impl}
To analyse the trade-off between accuracy and computational cost at test time, we evaluate both classical and deep learning based approaches. In a real world application, both of them can be considered depending on the scenario.

As a classical approach, we use the LOMO feature description and the XQDA metric learning algorithms \cite{LOMO}, as they aim at being effective and computationally efficient. As we discussed in section \ref{sec:SOA_prid}, LOMO presents the best trade-off between accuracy and computational cost for all the methods considered in the exhaustive analysis performed in \cite{karanam2016systematic}.

In a deep learning based approach, as described in section \ref{sec:SOA_prid}, the feature representations are extracted from a CNN considering the identities as classes and taking the output from the last layer before the softmax layer as the \textit{deep features}.
Our baseline is the one presented at \cite{PastPresentFuture} for the Market-1501 dataset, using the ResNet-50 \cite{Resnet} model. 
Since ResNet-50 might be too large for the datasets we consider, we also explore another smaller networks that can be more efficient and still perform well. 
In particular, we consider MobileNets \cite{mobilenets} as an alternative architecture.

MobileNets are presented as efficient light weight models suitable for mobile applications. 
The MobileNets architecture can be adapted to particular requirements of the system. In order to decide the network size, two parameters are introduced to control its latency and accuracy: the width multiplier $\alpha \in \left(0,1\right]$ and the resolution multiplier $\rho \in \left(0,1\right]$. The width multiplier can make the model thinner, by multiplying the number of input and output channels on each layer by $\alpha$.
$\rho$ is implicitly selected when determining the input size of the network, that can be 224, 192, 160 and 128.
Finally, as the similarity metric to compare the features extracted from the gallery images, we use the Euclidean distance.

\paragraph{Hand-crafted features}
To evaluate the LOMO features independently to XQDA, we compare the Euclidean distance, KISSME \cite{KISSME} and XQDA as similarity metrics.
PCA is commonly applied previously to KISSME in order to reduce the dimensionality of the LOMO features, in our case from 26960 to 200.
XQDA allows to select the dimensionality of its subspace. Thus, we also evaluate the performance of LOMO + XQDA depending on the XQDA dimensionality. The maximum value that we consider is the highest one with eigenvalues greater than 1. Following this criteria, we get a maximum dimensionality of 76 for the features extracted from the Market-1501 dataset. Therefore, we consider values of the XQDA dimensionality from 25 to 75.
Finally, to evaluate the computational cost, we measure the inference time of the method, running these experiments on a laptop with a CPU Intel Core i5-6300U CPU @ 2.40GHz.

\paragraph{Deep features}
Our deep learning based methods are implemented using the TensorFlow library.
The training and validation splits used for deep features are the ones provided on the original baselines. For Market-1501, \citet{PastPresentFuture} use a validation split of 1,294 images leaving 11,642 for training. The baseline for DukeMTMC-reID \cite{zheng2017unlabeled} uses the whole set of training images.
Finally, to evaluate the computational cost, we measure the inference time, running the experiments on a NVIDIA GTX1070 GPU. 

\begin{itemize}
  \item \textit{ResNet-50}\hspace{0.3cm}
  The ResNet-50 network is fine-tuned from the weights pre-trained for ImageNet, considering the person identities as classes. The deep features are then extracted from the last layer before the softmax, which in the ResNet-50 architecture, corresponds to the output of the average pooling layer.
  
  It is worth mentioning that because of the high number of classes in the datasets (751 and 702 identities for the training splits of Market-1501 and DukeMTMC-reID respectively), with few images per class (a mean of 20 for DukeMTMC-reID and 17.2 for Market-1501), it is hard to train the network, since a deep neural network needs a big amount of data to converge properly.
  
  To train ResNet-50, we resize the input images to 224x224 and use horizontal flip for data augmentation. Using Stochastic Gradient Descent (SGD), we initially set the learning rate to 0.001 with a decay of 0.1 every 20000 steps. Using a batch size of 16 and momentum of 0.9, we train the network for 21 epochs (15000 iterations) for the Market-1501 dataset. For DukeMTMC-reID, the learning rate is initially set to 0.01 and we use a batch size of 32, training it for 29 epoch (15000 iterations).
  
  \item \textit{MobileNets}\hspace{0.3cm} 
  We choose an input size of 128 due to the size of the images of the datasets we use. Market-1501 images have a fixed size of 128x64 while DukeMTMC-reID images size vary. Then, we resize all the images to 128x128, applying horizontal flip for data augmentation.
  We evaluate the performance for width multipliers of $\alpha=0.25, 0.5, 0.75, 1.0$, which are the values with ImageNet pre-trained weights being provided. We denote these networks as MobileNet 0.25, 0.5, 0.75 and 1.0 respectively. $\alpha$ also affects the dimensionality of the extracted features from the network, which are the output of the final average pooling. The features are of length 1024, 768, 512 and 256 for values of $\alpha=1.0, 0.75, 0.5$ and 0.25 respectively.
  
  The training hyperparameters we use, are those that perform best across several experiments. 
  The results on Market-1501 are obtained by using SGD with a batch size of 32, an initial learning rate of 0.01 with a decay of 0.1 every 20000 steps, and momentum of 0.9. We train MobileNet 0.25 for 29 epochs and the rest of MobileNets for 39 epochs.
  On DukeMTMC-reID, we set the initial learning rate to 0.01 for MobileNet 0.25 and to 0.02 for MobileNet 1.0, training both of them for 39 epochs. For MobileNets 0.5 and 0.75, we use a batch size of 16, a starting learning rate of 0.005 and we train them for 39 epochs.
\end{itemize}

\begin{figure}[H]
\begin{minipage}{\linewidth}
  \centering
  \centerline{\includegraphics[scale = 0.3]{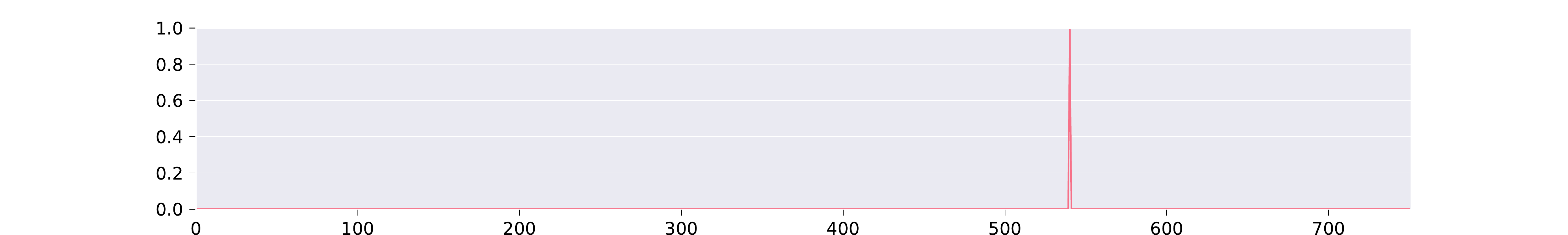}}
\end{minipage}
\vfill
\begin{minipage}{\linewidth}
  \centering
  \centerline{\includegraphics[scale = 0.3]{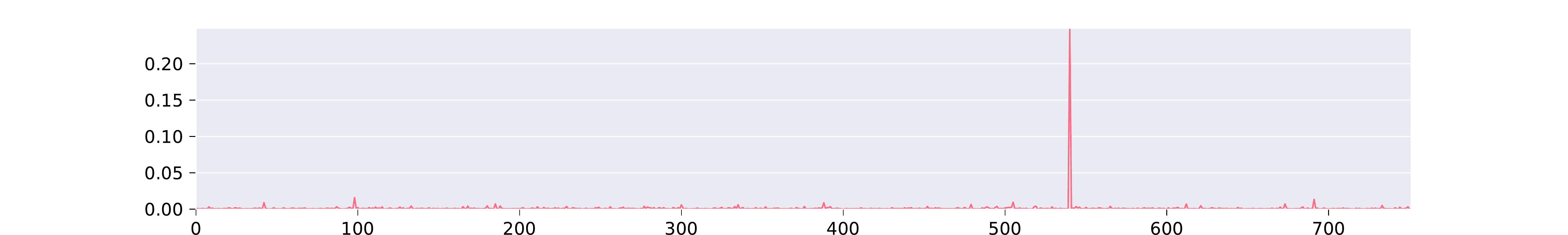}}
\end{minipage}\vfill
\begin{minipage}{\linewidth}
  \centering
  \centerline{\includegraphics[scale = 0.3]{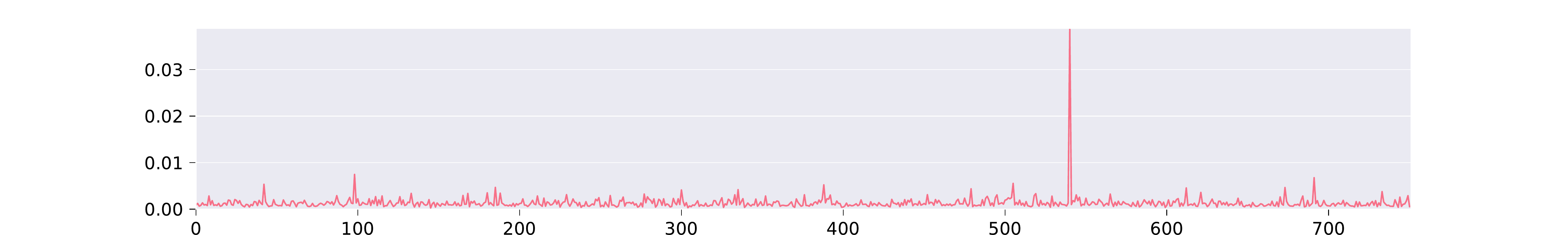}}
\end{minipage}
\begin{minipage}{\linewidth}
  \centering
  \centerline{\includegraphics[scale = 0.3]{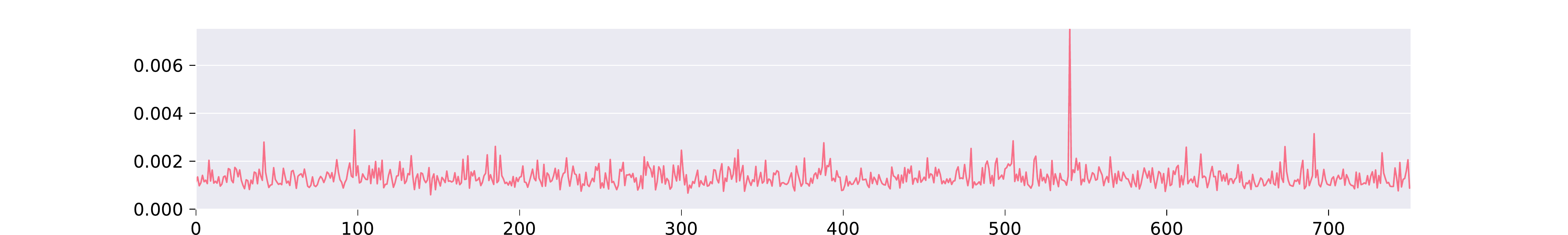}}
\end{minipage}
\begin{minipage}{\linewidth}
  \centering
  \centerline{\includegraphics[scale = 0.3]{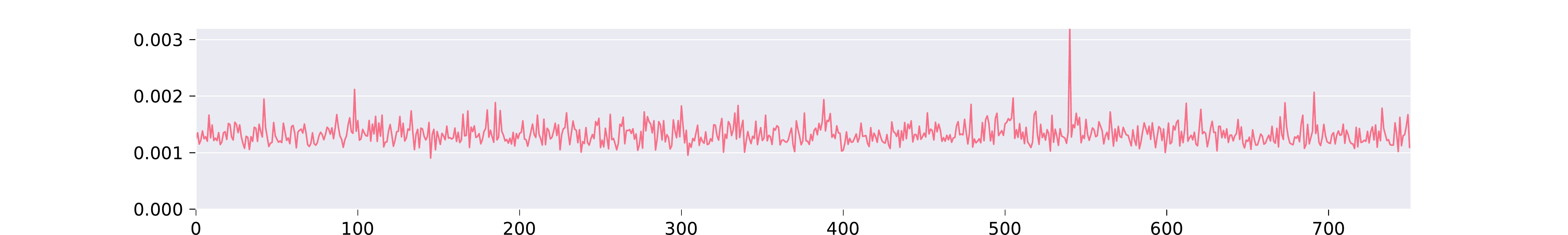}}
\end{minipage}
\begin{minipage}{\linewidth}
  \centering
  \centerline{\includegraphics[scale = 0.3]{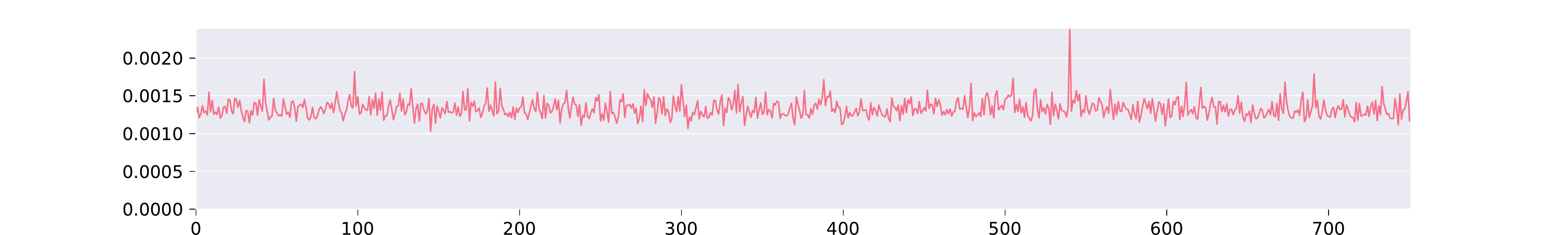}}
\end{minipage}
\caption{Original and softened probability distributions generated by the teacher network for temperature values of (from top to bottom) T=1,3,5,10,20,30.}
\label{fig:softened_dist}
\end{figure}

\begin{figure}[H]
\begin{minipage}{\linewidth}
  \centering
  \centerline{\includegraphics[scale = 0.47]{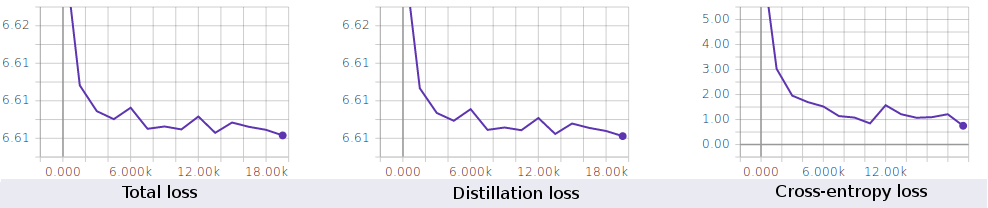}}
  \centerline{(a)}\medskip
\end{minipage}
\vfill
\begin{minipage}{\linewidth}
  \centering
  \centerline{\includegraphics[scale = 0.47]{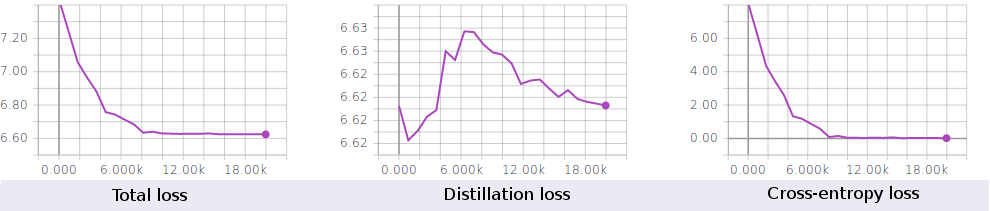}}
  \centerline{(b)}\medskip
\end{minipage}
\caption{Training loss for the distillation with (a) low ($\lambda=0.0001$) and (b) high ($\lambda=0.1$) $\lambda$ values. The distillation loss leads the training in case (a), while in case (b) it is done by the cross-entropy loss (see equation \ref{eq:lossdistill}).}
\label{fig:distill_loss}
\end{figure}

\paragraph{Network distillation}
We propose ResNet-50 as teacher, but also MobileNet 1.0, which has the biggest capacity among the MobileNets configurations. The number of parameters for MobileNets are 4.24M, 2.59M, 1.34M and 0.47M for width multiplier values of 1.0, 0.75, 0.5 and 0.25 respectively, while ResNet-50 has 23.5M of parameters.
Since we want an efficient network, the student is the MobileNet with the smallest width multiplier (MobileNet 0.25).
We analyse the effect of the hyperparameters of the distillation, that are the temperature T and the regularization weight $\lambda$ for the distillation loss. We consider the range of temperatures $1-30$, being $T=1$ the case in which the entropy of the soft targets is not modified and $T=30$ a case of very high temperature. The highest temperature is selected based on the observed softened probability distribution that is generated by the teacher network for $T=30$, as it is shown in Fig. \ref{fig:softened_dist}. In that probability distribution, the difference between the probabilities assigned to the incorrect classes and the one assigned to the correct class is less than a $0.1\%$. This is due to a very high temperature with which the probability distribution is almost flat (which is the case of maximum entropy).
To do the analysis for T in that range, we use intervals of 5, and 1 for the lowest values. For $\lambda$, we choose the values 0.0001, 0.001 and 0.01. They have been chosen by analysing the contribution of the loss terms while monitoring the training process, as shown in Fig. \ref{fig:distill_loss}. When using a value of $\lambda=0.1$, the cross-entropy loss leads the training and the distillation term barely affects, but we noted from our experiments that it makes the training harder to converge, resulting in a performance drop. Therefore, we do not consider $\lambda=0.1$ and higher values for our analysis.

For each value of T, we evaluate both the Rank-1 accuracy and mAP with the features extracted from the student network. We try several combinations of the hyperparameters, \textit{i.e.}, learning rate, batch size, number of epochs, etc. However, most of the experiments perform best using the same hyperparameters, \textit{i.e}, we obtain that the same optimum configuration of parameters for several values of T and $\lambda$.
Then, all the Rank-1 and mAP values reported in section \ref{sec:results} for each value of T, are those that perform best among all the experiments performed. Most of the distillation experiments use SGD, with an initial learning rate of 0.02 that decays 0.1 every 20000 steps, and a momentum of 0.9, being trained for 39 epochs.

\section{Results}
\label{sec:results}

The performance of the classical approach using LOMO and XQDA is shown in Table \ref{table:LOMOXQDAresults}. We verify that the usage of metric learning algorithms such as KISSME or XQDA significantly improves the performance of hand-crafted features. However, we must consider  that in this table, PCA is previously applied in the case of KISSME to reduce the dimensionality of the LOMO features to 200. The dimensionality in the XQDA space is 75, which is considerably smaller. Thus, XQDA performs better than KISSME even with a stronger dimensionality reduction.

However, both XQDA and KISSME require a metric learning step that increases the computational cost. In particular, the XQDA training, \textit{i.e} finding the projection matrix from the training set samples, takes 892 seconds for Market-1501, whose training set contains 12936 images.
Also, comparing a query image against the gallery takes a mean time of 1,951 ms per image. Thus, using XQDA, the system compares the individuals' features at a rate of $0.5$ images$/$s.
Regarding the computational cost for feature extraction with LOMO, the mean CPU time to extract the LOMO features per image is 17.5ms. Then, the system is able to get the descriptors for the images of the individuals at a rate of 57 images$/$s.

\begin{table}[h]
\caption{LOMO and XQDA performance on Market-1501.}
\begin{center}
\begin{tabular}{c c c c}
\hline
\textbf{Features} & \textbf{Similarity metric} & \textbf{Rank-1} (\%) & \textbf{mAP} (\%)\\
\hline
LOMO & Euclidean distance & 27.11 & 8.01\\

LOMO & KISSME \cite{KISSME} & 41.83	& 19.37\\

LOMO & XQDA (dimensionality 75) & 43.32	& 22.01\\
\hline
\end{tabular}
\end{center}
\label{table:LOMOXQDAresults}
\end{table}

\begin{figure}[h]
\begin{minipage}{0.45\linewidth}
  \centering
  \centerline{\includegraphics[scale = 0.45]{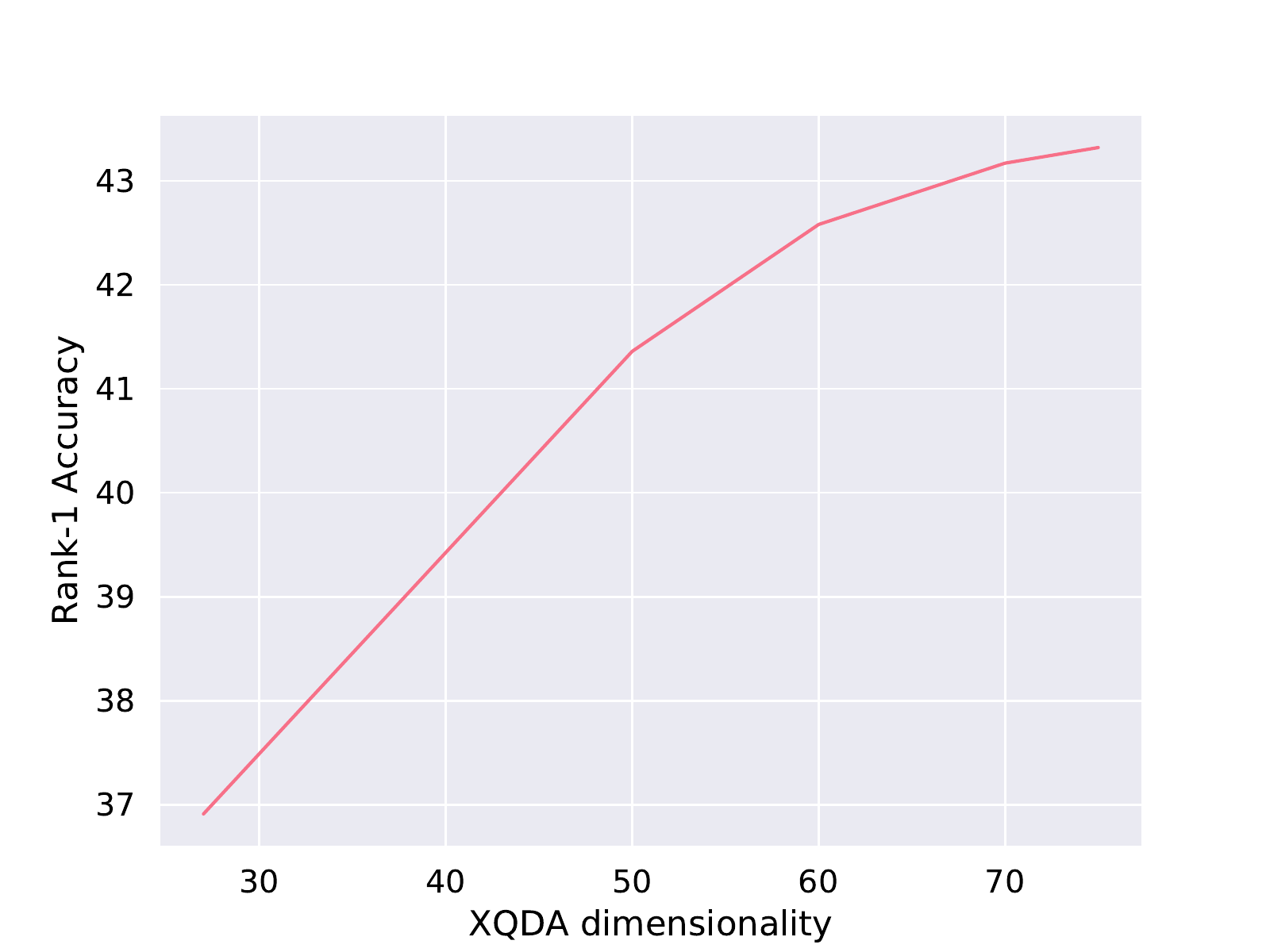}}
  \centerline{(a)}
\end{minipage}
\hfill
\begin{minipage}{0.40\linewidth}
  \centering
  \centerline{\includegraphics[scale = 0.47]{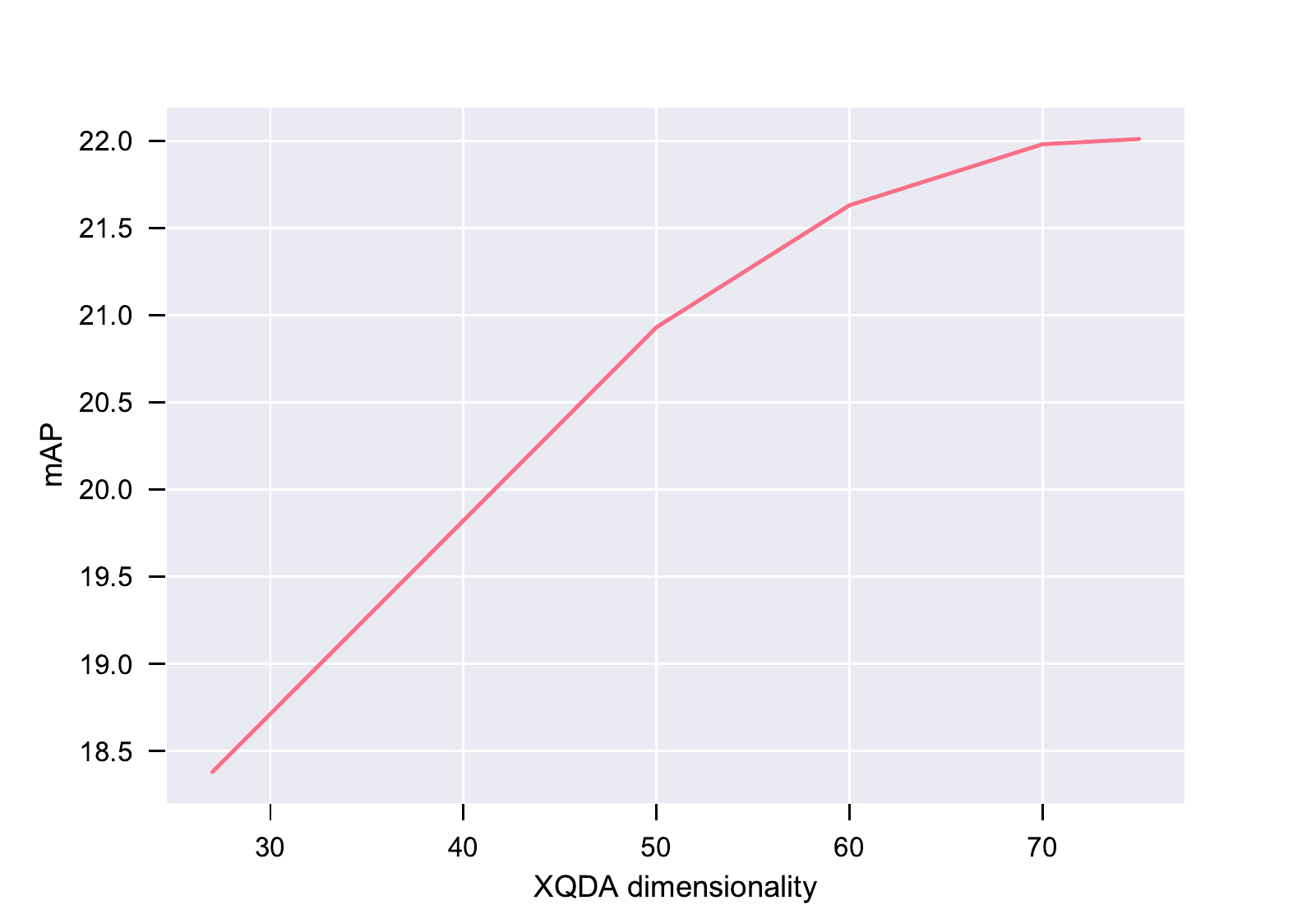}}
  \centerline{(b)}
\end{minipage}
\caption{Performance for LOMO + XQDA on Market-1501 depending on the XQDA dimensionality. (a) Rank-1 accuracy and (b) Mean average precision.}
\label{fig:XQDAdim}
\end{figure}

The performance of LOMO+XQDA reported in Table \ref{table:LOMOXQDAresults} corresponds to the highest dimensionality value for XQDA. We also show the dependency of the performance with the XQDA dimensionality on Fig. \ref{fig:XQDAdim}. The accuracy increases with the dimensionality of XQDA, since more information can be encoded in the feature vector with a higher dimensionality. Although we expect a saturation on the performance from a certain value, we do not reach such value. This is probably because the maximum dimensionality in our case is 75, which is considerably low. It is much lower than the dimensionality of the smallest feature vectors considered in this work that is 256 for MobileNet 0.25.

For the deep features baseline, \citet{PastPresentFuture} get a 72.54\% of rank-1 accuracy and 46\% mean average precision on the Market-1501 dataset, with deep features extracted from ResNet-50. Following the same strategy, in \cite{zheng2017unlabeled} the baseline results for the DukeMTMC-reID dataset are a 65.22\% of rank-1 accuracy and 44.99\% of mean average precision.

Fine-tuning the ResNet-50 and MobileNets architectures to the datasets considered, we obtain the performance presented in Table \ref{table:finetuning_results}. 
For Market-1501, the middle size MobileNets are the models that perform best, even slightly better than the biggest one and ResNet-50. However, MobileNet 0.25 presents a lower performance.
The reason why the middle models perform so well, could be that all of them have enough capacity to solve the problem. Then, a bigger architecture, such as ResNet-50, would not involve an improvement.
Moreover, as mentioned in section \ref{sec:Exp_impl}, training the networks on a dataset with a high number of classes and a small number of samples per class is not straightforward. The baseline achieved with ResNet-50 by \citet{PastPresentFuture} suggests that a higher performance could be achieved for this network.

For the DukeMTMC-reID dataset, MobileNets do not perform as good as they do for Market-1501. 
The reason might be that this dataset is more challenging, and requires a higher capacity of the network to perform a good description of the identities. Since the size of the bounding boxes vary and all of them have to be resized to 128x128, losing thereby the aspect ratio, the input images have a higher variability.

\begin{table}[h]
\caption{Rank-1 accuracy, mean Average Precision (mAP) and computational cost of the inference for the deep features from the ResNet-50 and MobileNet architectures trained on the Market-1501 and DukeMTMC-reID datasets.}
\begin{center}
\begin{tabular}{c c c c}
\hline
\textbf{Market-1501} & \textbf{Rank-1} (\%) & \textbf{mAP} (\%) & \textbf{\# images/s} \\
\hline
ResNet-50 & 64.46 & 38.95 & 128 \\ 
MobileNet 0.25 & 59.74 & 34.13 & 613 \\ 
MobileNet 0.5  & 68.11 & 41.52 & 607 \\ 
MobileNet 0.75  & 67.34 & 40.44 & 574 \\ 
MobileNet 1.0 & 67.37 & 39.54 & 545 \\ 
\hline
\textbf{DukeMTMC-reID} & \textbf{Rank-1} (\%) & \textbf{mAP} (\%)& \textbf{\# images/s}\\
\hline
ResNet-50 & 67.1 & 44.59 & 128\\
MobileNet 0.25 & 49.69 & 28.67 & 613 \\
MobileNet 0.5 & 54.62 & 32.17 & 607 \\
MobileNet 0.75 & 57.32 & 34.69 & 574\\
MobileNet 1.0 & 57.41 & 34.86 & 545 \\
\hline
\end{tabular}
\end{center}
\label{table:finetuning_results}
\end{table}

\begin{figure}[H]
\begin{minipage}{.4\linewidth}
  \centering
  \centerline{\includegraphics[scale = 0.48]{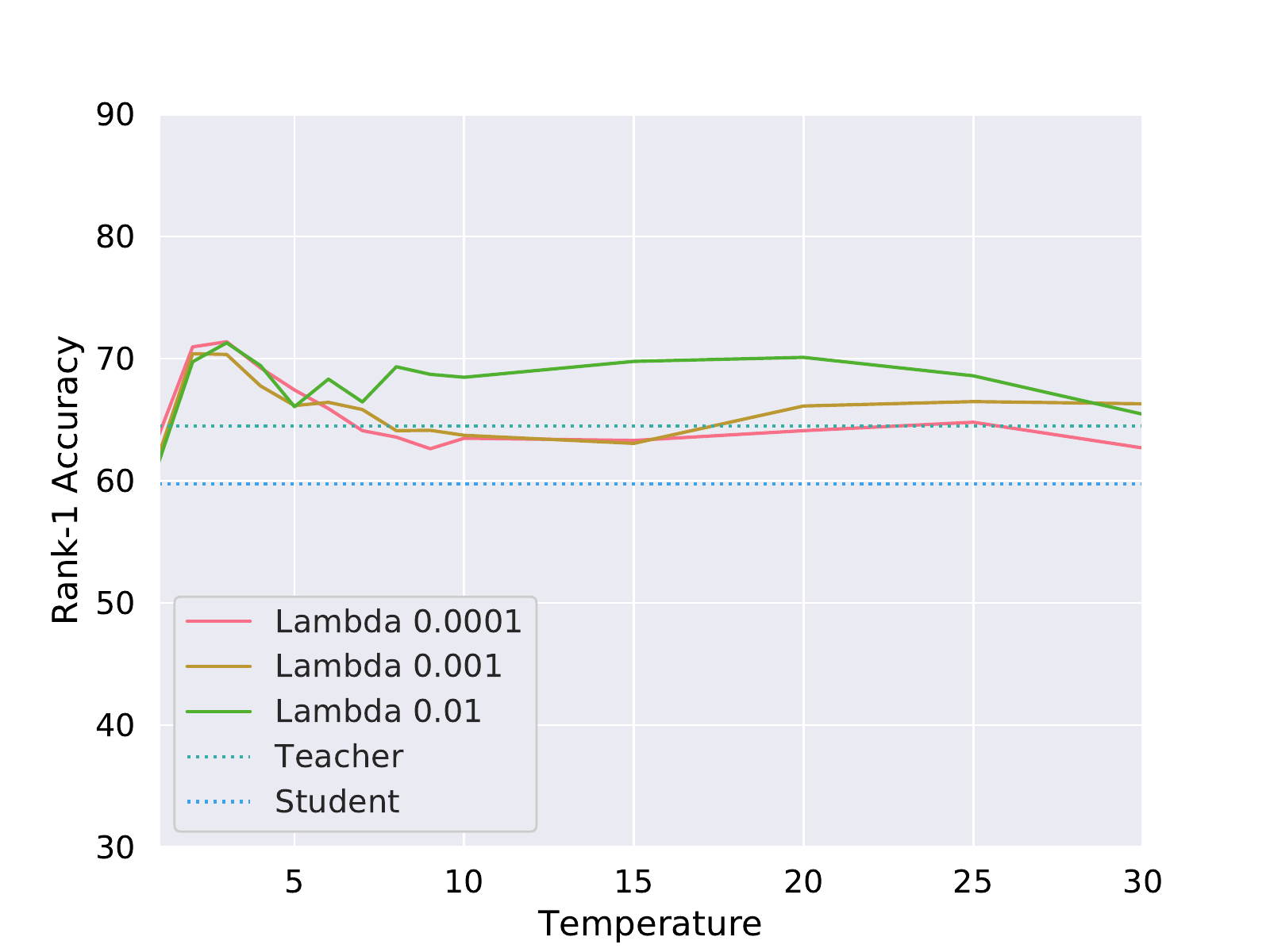}}
  \centerline{(a)}
\end{minipage}
\hfill
\begin{minipage}{0.4\linewidth}
  \centering
  \centerline{\includegraphics[scale = 0.48]{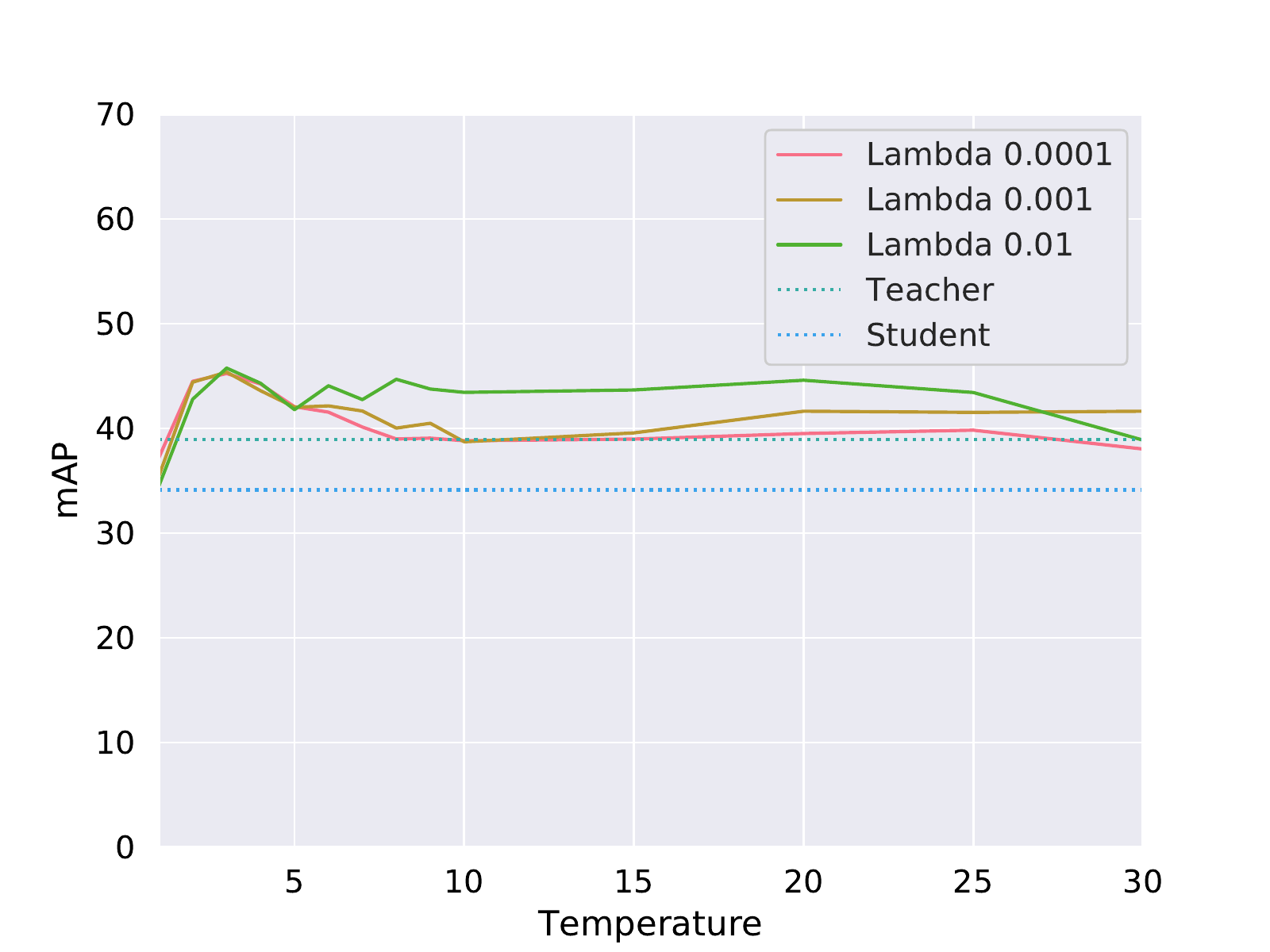}}
  \centerline{(b)}
\end{minipage}

\begin{minipage}{.4\linewidth}
  \centering
  \centerline{\includegraphics[scale = 0.48]{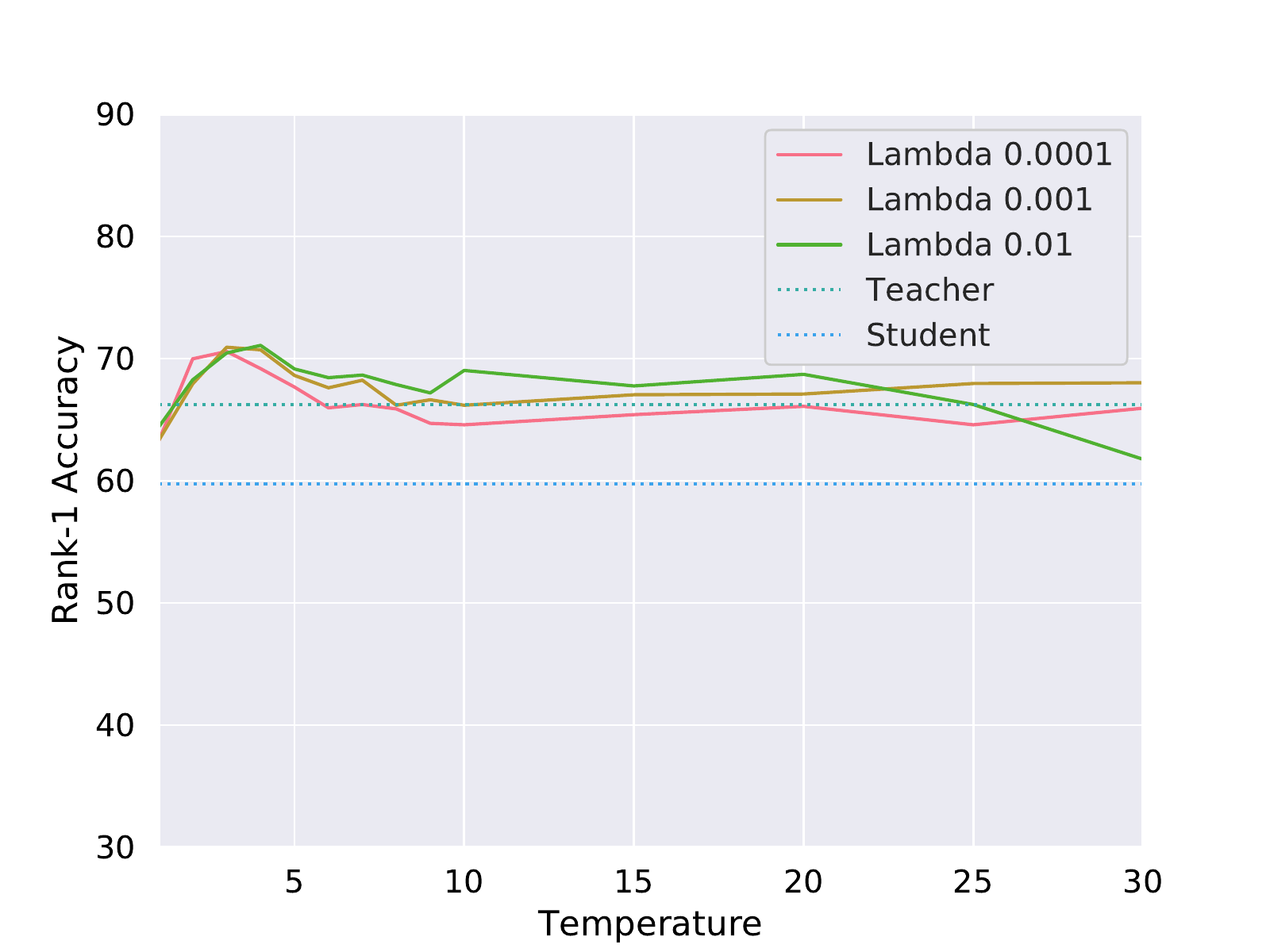}}
  \centerline{(c)}
\end{minipage}
\hfill
\begin{minipage}{0.4\linewidth}
  \centering
  \centerline{\includegraphics[scale = 0.48]{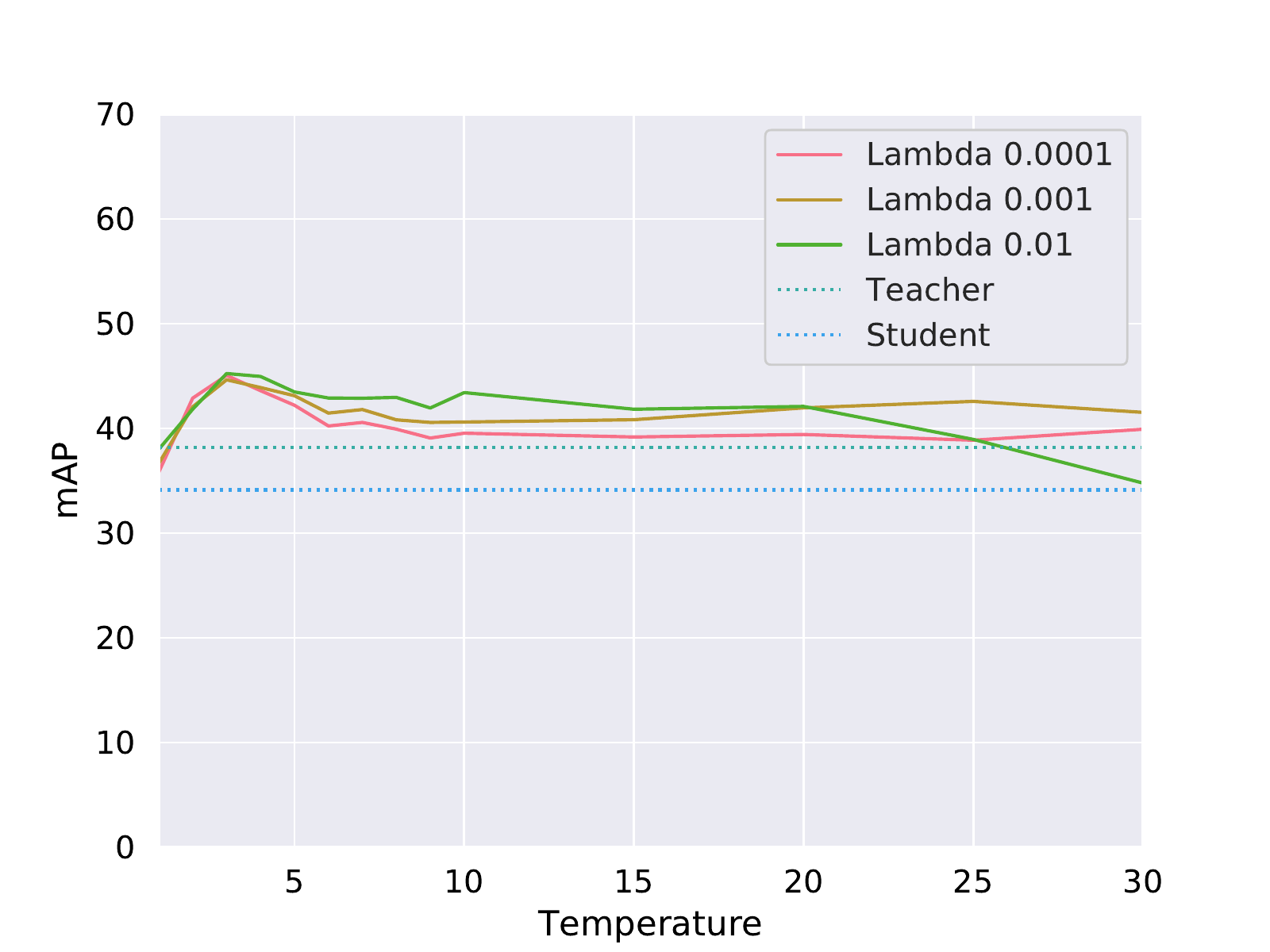}}
  \centerline{(d)}
\end{minipage}
\caption{Distillation performance on Market-1501. (a) Rank-1 accuracy and (b) Mean average precision for student model MobileNet 0.25 with teacher model ResNet-50.  (c) Rank-1 accuracy and (d) Mean average precision for student model MobileNet 0.25 with teacher model MobileNet 1.0. Best viewed in colour.}
\label{fig:marketdistil}
\end{figure}

\begin{figure}[H]
\begin{minipage}{.4\linewidth}
  \centering
  \centerline{\includegraphics[scale = 0.48]{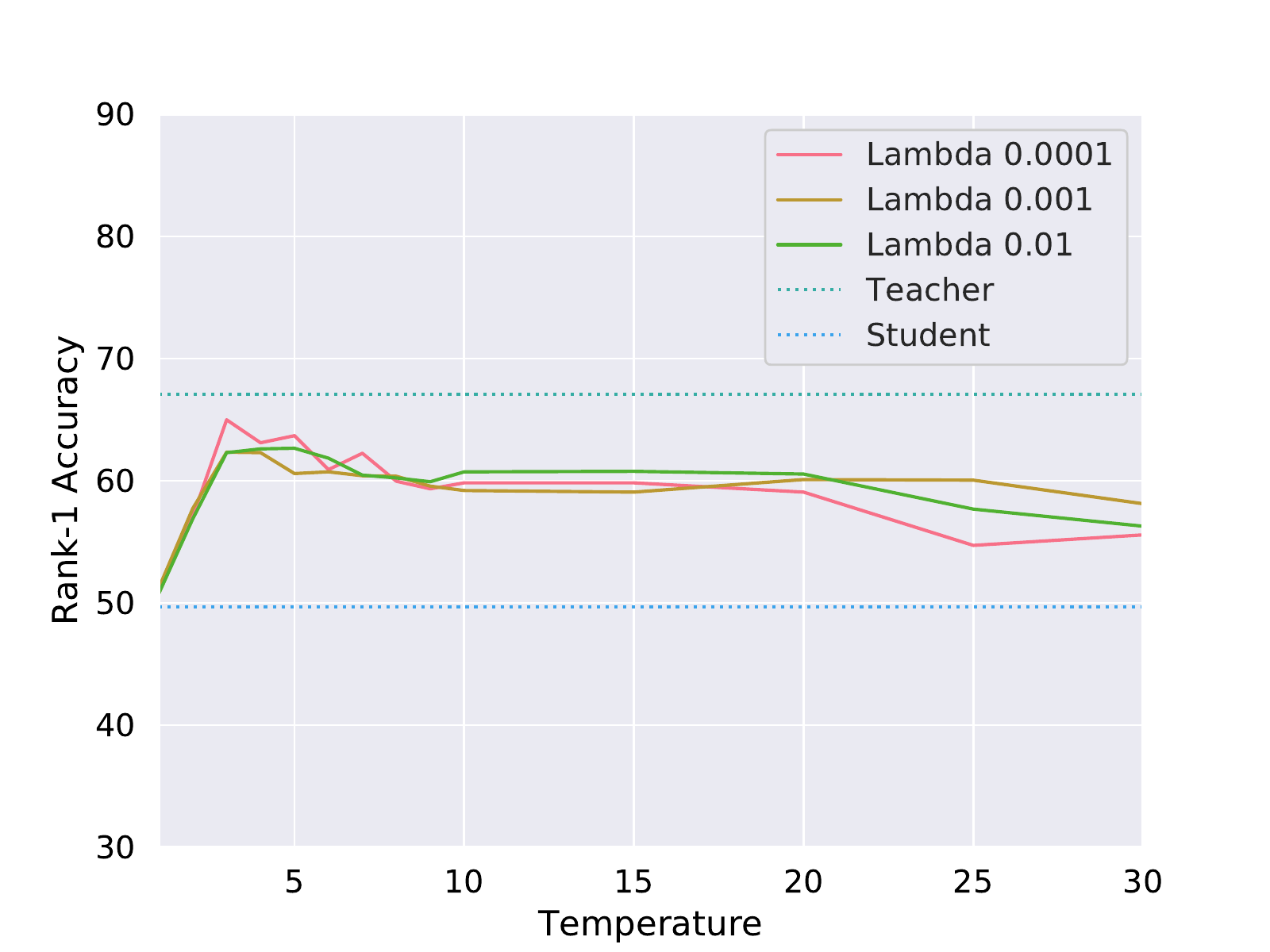}}
  \centerline{(a)}
\end{minipage}
\hfill
\begin{minipage}{0.4\linewidth}
  \centering
  \centerline{\includegraphics[scale = 0.48]{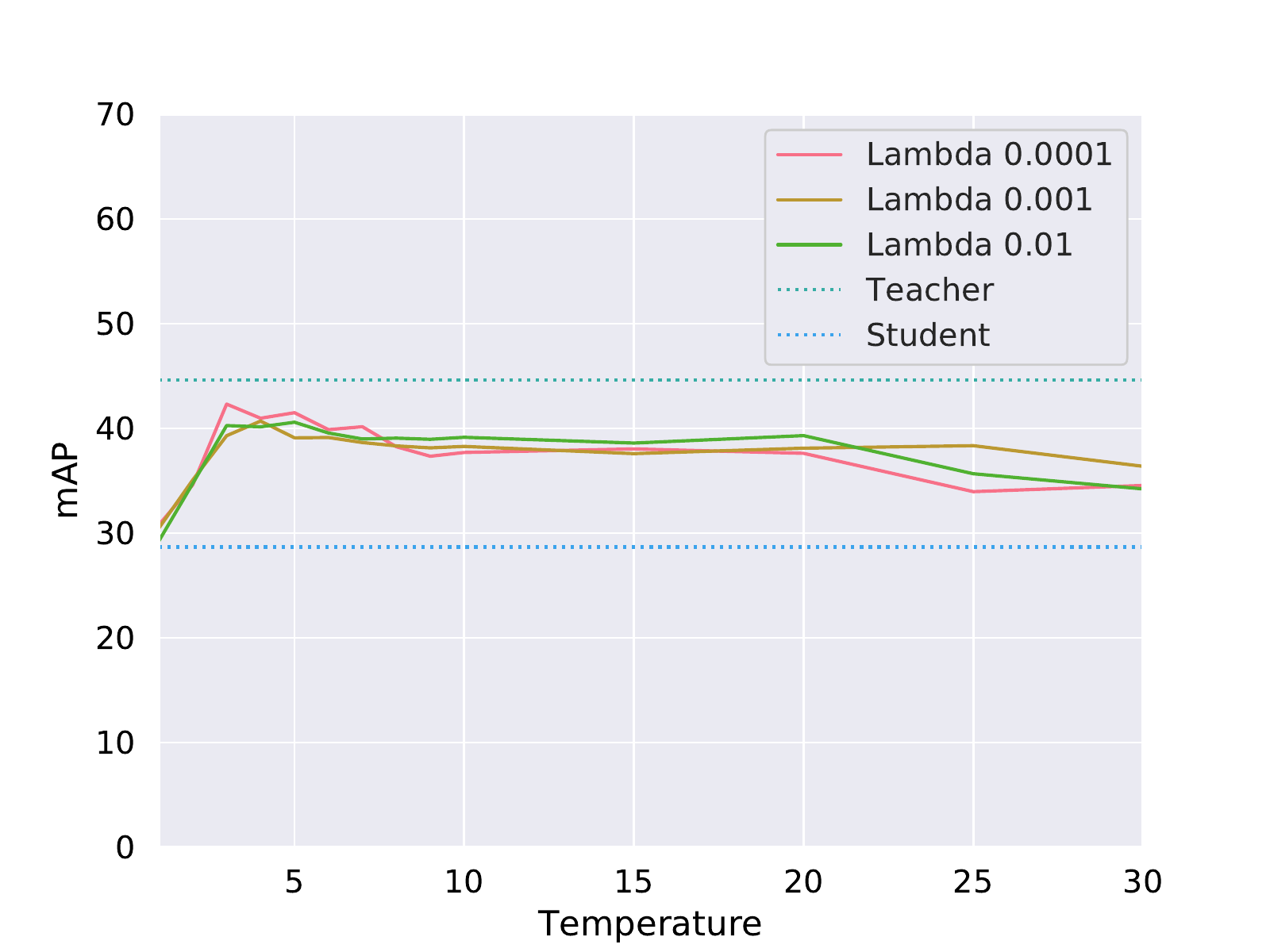}}
  \centerline{(b)}
\end{minipage}

\begin{minipage}{.4\linewidth}
  \centering
  \centerline{\includegraphics[scale = 0.48]{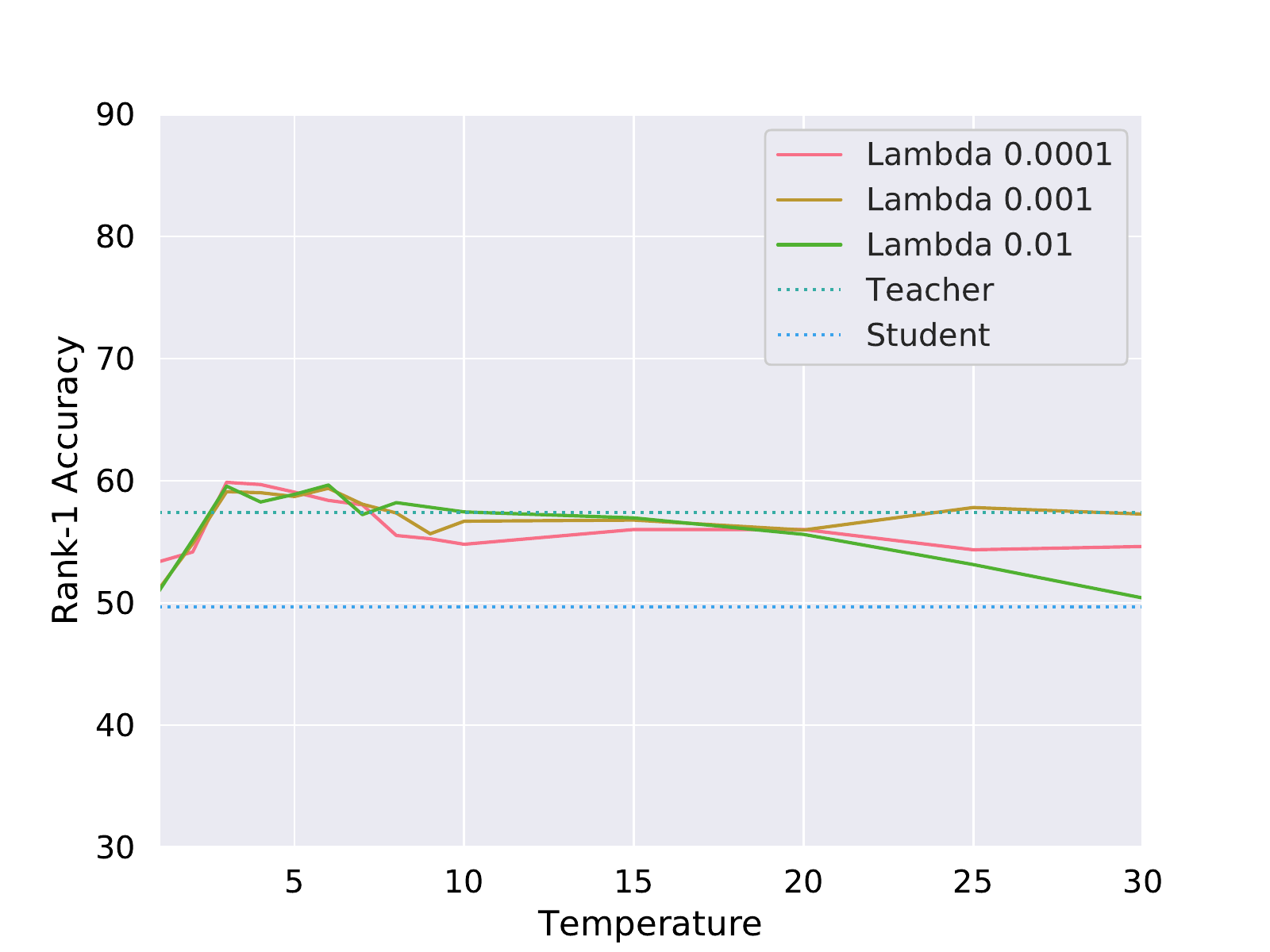}}
  \centerline{(c)}
\end{minipage}
\hfill
\begin{minipage}{0.4\linewidth}
  \centering
  \centerline{\includegraphics[scale = 0.48]{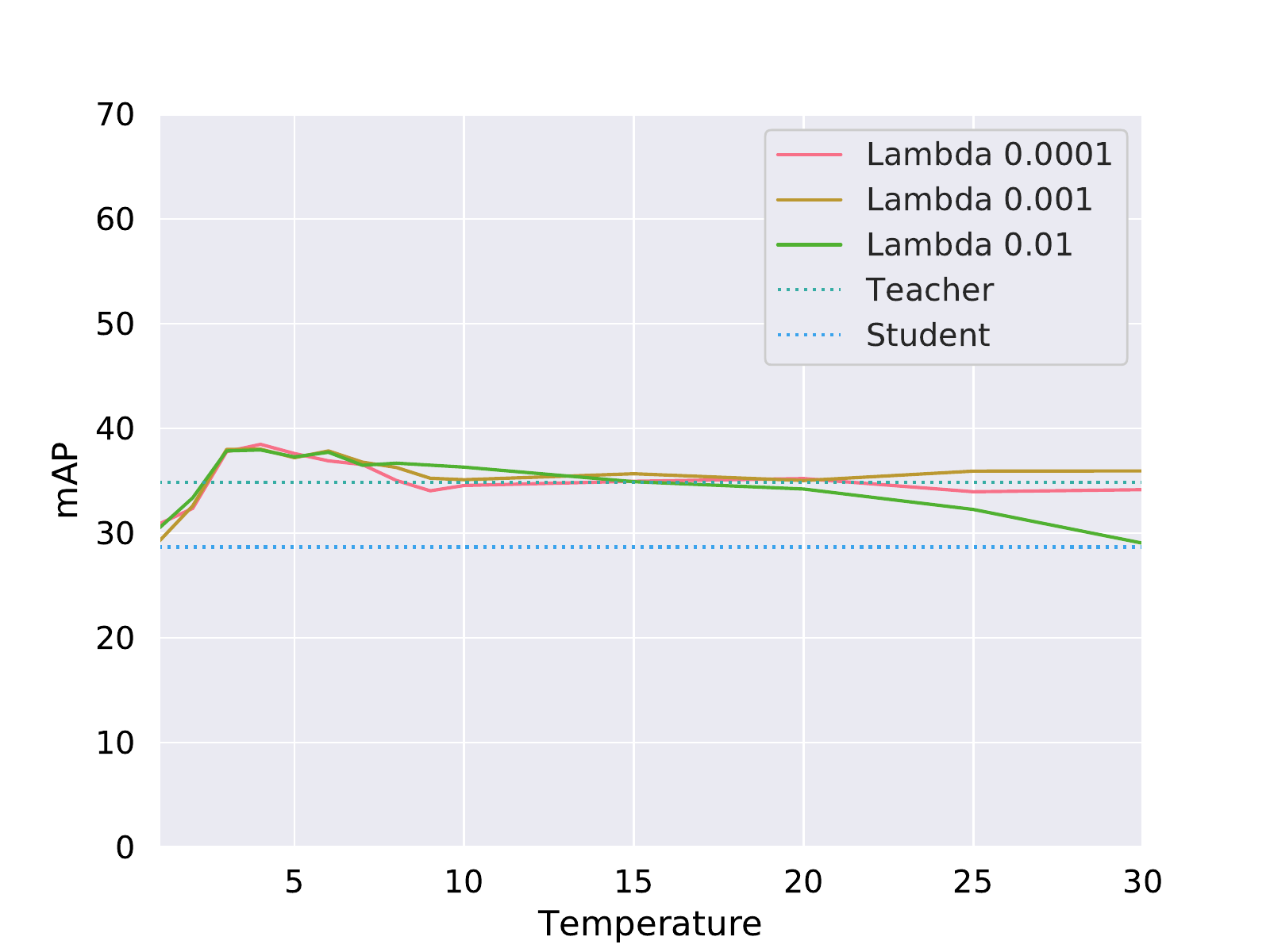}}
  \centerline{(d)}
\end{minipage}

\caption{Distillation performance on DukeMTMC-reID. (a) Rank-1 accuracy and (b) Mean average precision for student model MobileNet 0.25 with teacher model ResNet-50.  (c) Rank-1 accuracy and (d) Mean average precision for student model MobileNet 0.25 with teacher model MobileNet 1.0. Best viewed in colour.}
\label{fig:dukedistil}
\end{figure}

We perform the network distillation experiments using pre-trained ResNet-50 and MobileNet 1.0 networks as teachers, whose performance is reported in Table \ref{table:finetuning_results}.
We show in Fig. \ref{fig:marketdistil} and Fig. \ref{fig:dukedistil} the Rank-1 accuracy and mAP dependency with the temperature in the distillation, for the Market-1501 and DukeMTMC-reID datasets respectively.
The performance of the teacher and the student trained independently is also drawn in the previous figures to provide the comparison with the baseline without distillation. All the experiments improve significantly the performance of the student, and even the performance of the teacher for low temperatures. 
The only case in which the student does not outperform the teacher is for the DukeMTMC-reID dataset for the distillation from ResNet-50 (Fig. \ref{fig:dukedistil} (a,b)). However, in this case, the difference of performance between the teacher and the student is higher than for the other experiments.

For a fixed value of $\lambda$, there is always a peak of performance in $T=3$. The worst performance across all the values of the temperature T, is for T=1, which corresponds to the case in which the temperature is not increased, \textit{i.e.} the original logits from the teacher models are used. This demonstrates the importance of raising the temperature to produce suitable soft targets. Also, from a certain value of T, the performance gets saturated, probably because the probabilities are already very softened and they do not change significantly for those values of T, as Fig. \ref{fig:softened_dist} (e,f) shows for the values of $T=20, 30$. The differences of probabilities among both distributions are less than a $0.1\%$.

\begin{figure}[H]
\begin{minipage}{.4\linewidth}
  \centering
  \centerline{\includegraphics[scale = 0.5]{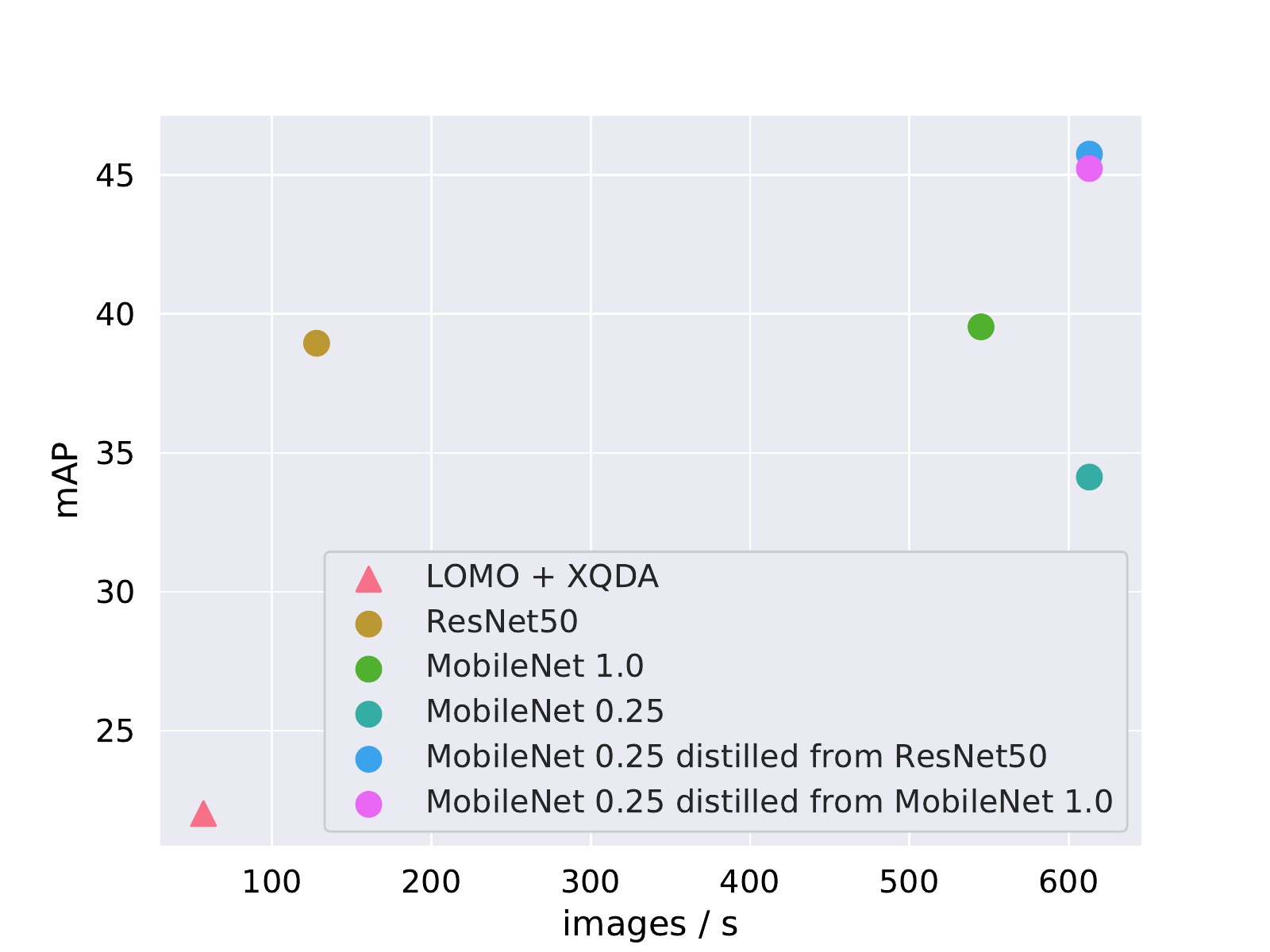}}
  \centerline{(a)}
\end{minipage}
\hfill
\begin{minipage}{0.4\linewidth}
  \centering
  \centerline{\includegraphics[scale = 0.5]{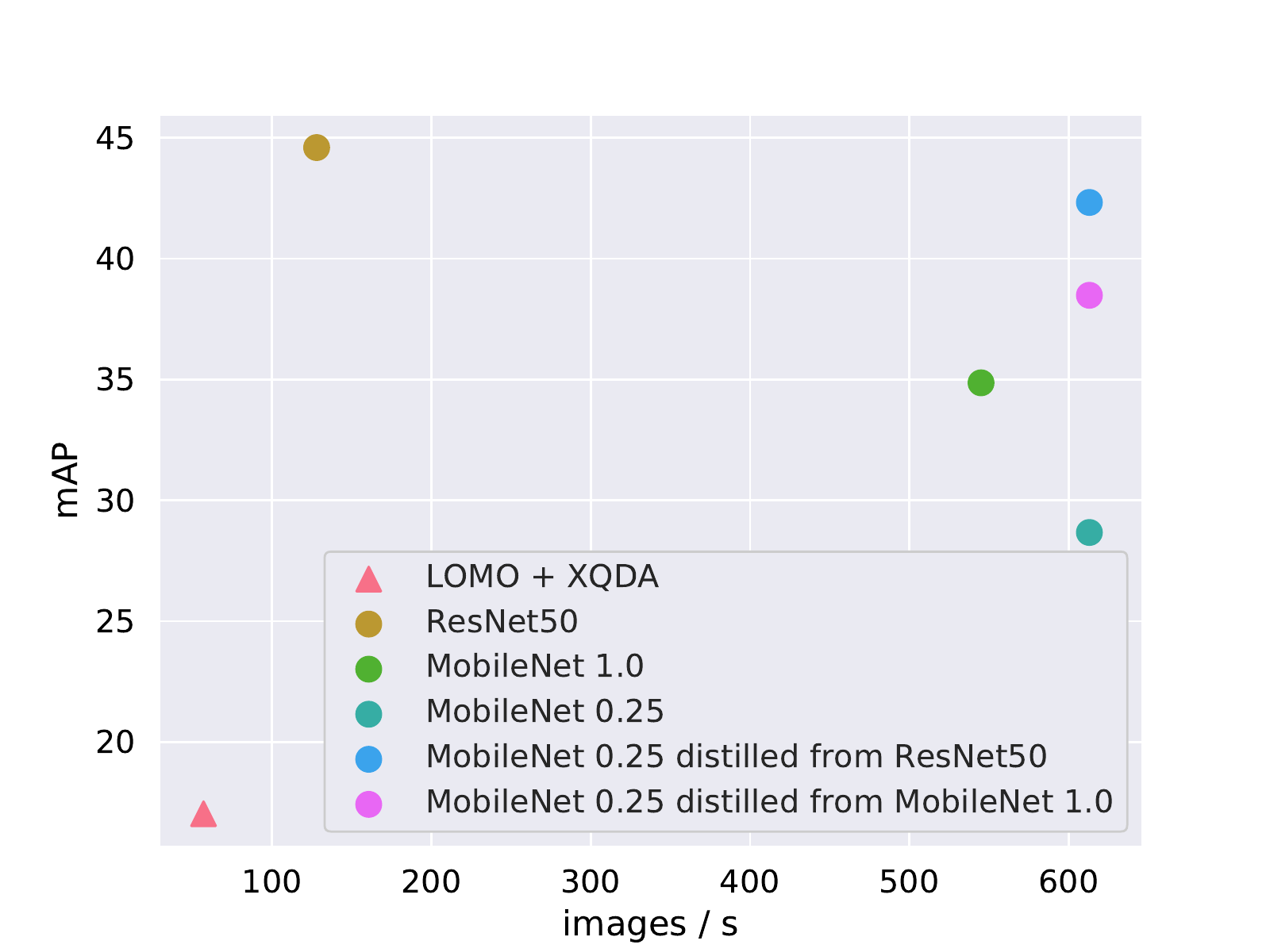}}
  \centerline{(b)}
\end{minipage}
\caption{Trade-off between the mean average precision (mAP) and the feature extraction time for the proposed methods on the (a) Market-1501 and (b)  DukeMTMC-reID datasets. Note that the feature extraction time for LOMO is measured as CPU time while the deep features experiments are run on a GPU. Best viewed in colour.}
\label{fig:tradeoff}
\end{figure}

In Table \ref{table:SOAcomparison}, we compare our configuration with the highest performance for network distillation against the state-of-the-art. Although the accuracy achieved is not better than the state-of-the-art, our method is specifically designed to be efficient, which can compromise the accuracy.

Finally, to summarise all the considered methods, we show in Fig. \ref{fig:tradeoff} and in Table \ref{table:tradeoff}, the trade-off between computational cost and accuracy. 
In this table, we compare the performance of the classical approach (LOMO+XQDA), the deep features extracted from the MobileNets architectures trained with the cross-entropy loss as well as the deep features extracted from MobileNet 0.25 being distilled from the MobileNet 1.0 and ResNet-50 models, whose performance is reported in the table. On the Market-1501 dataset, we compute the LOMO features and then apply XQDA with dimensionality 75, while the results for the DukeMTMC-reID dataset is from \cite{zheng2017unlabeled}. 

\begin{table}[H]
\caption{Rank-1 accuracy and mean Average Precision (mAP) for network distillation, taking MobileNet ($\alpha=$0.25) as the student network, and MobileNet ($\alpha=$1.0) and ResNet-50 as the teachers, compared against the state-of-the-art on the Market-1501 and DukeMTMC-reID benchmarks.}
\begin{center}
\begin{tabular}{c c c}
\hline
\textbf{Market-1501} & \textbf{Rank-1} (\%) & \textbf{mAP} (\%)\\
\hline
MobileNet 0.25 distilled from ResNet-50 & 71.29	& 45.76\\
MobileNet 0.25 distilled from MobileNet 1.0 & 70.46	& 45.24\\
\hline
P2S \cite{P2S} & 70.72 & 44.27\\
CADL \cite{Lin_2017_CVPR} & 73.84 & 47.11 \\
MSCAN Fusion \cite{Li_2017_CVPR} & 80.31 & 57.53 \\
SVDNet \cite{sun2017svdnet} & 82.3 & 62.1 \\
ACRN \cite{schumann2017person}  & 83.61 & 62.60\\
DML \cite{lu2018cvpr} & 89.34 & 70.51 \\
FD-GAN \cite{ge2018nips} & 90.5 & 77.7 \\
\hline
\textbf{DukeMTMC-reID} & \textbf{Rank-1} (\%) & \textbf{mAP} (\%) \\
\hline
MobileNet 0.25 distilled from ResNet-50 & 64.99 &	42.32\\
MobileNet 0.25 distilled from MobileNet 1.0 & 59.69 & 38.48\\
\hline
Dataset baseline with ResNet-50 \cite{zheng2017unlabeled} & 65.22 & 44.99 \\
ACRN \cite{schumann2017person} & 72.58 & 51.96\\
SVDNet \cite{sun2017svdnet} & 76.7 & 56.8 \\
FD-GAN \cite{ge2018nips} & 80.0 & 64.5 \\
\end{tabular}
\end{center}
\label{table:SOAcomparison}
\end{table}

Note that LOMO is measured in CPU time, while all the deep features methods are measured in GPU time. Therefore, the comparison for computational cost is not strictly fair. In terms of accuracy, the LOMO+XQDA accuracy is with a large margin the lowest, as expected for a hand-crafted method. Then, this kind of method would be suitable only for an application in which either a GPU, or a large amount of annotated data, is not available. 
The results show that distillation improves effectively the performance of efficient networks, providing the best accuracy among all the considered methods, as well as the lowest inference time. 
It is also worth mentioning the gap of computational cost between ResNet-50 and MobileNets, while their performance in terms of accuracy is very similar. Then, it is important to choose a suitable architecture for the problem we want to solve. For the Market-1501 dataset, a network of the size of MobileNet can describe the features of the identities effectively. In the case of DukeMTMC-reID, ResNet-50 performs much better.

\begin{table}[H]
\caption{Evaluation of the trade-off between Rank-1 accuracy, mean Average Precision (mAP) and computational time on the Market-1501 and DukeMTMC-reID datasets. \textit{d.f.} stands for "distilled from".}
\begin{center}
\begin{tabular}{c c c c}
\hline
\textbf{Market-1501} & \textbf{Rank-1} (\%) & \textbf{mAP} (\%) & \textbf{\# images/s} \\
\hline
LOMO + XQDA & 43.32 & 22.01 & 57\\
ResNet-50 & 64.46 & 38.95 & 128 \\
MobileNet 1.0 independent & 67.37 & 39.54 & 545 \\
MobileNet 0.25 independent & 59.74 & 34.13 & 613 \\
MobileNet 0.25 d.f. ResNet-50 & 71.29	& 45.76 & 613 \\
MobileNet 0.25 d.f. MobileNet 1.0 & 70.46	& 45.24 & 613\\
\hline
\textbf{DukeMTMC-reID} & \textbf{Rank-1} (\%) & \textbf{mAP} (\%)& \textbf{\# images/s}\\
\hline
 LOMO + XQDA \cite{zheng2017unlabeled} &  30.75 & 17.04 & 57\\
 ResNet-50 & 67.1 & 44.59 & 128\\
 MobileNet 1.0 independent & 57.41 & 34.86 & 545 \\
 MobileNet 0.25 independent & 49.69 & 28.67 & 613 \\
 MobileNet 0.25 d.f. ResNet-50 & 64.99 &	42.32 & 613\\
 MobileNet 0.25 d.f. MobileNet 1.0 & 59.69 & 38.48 & 613\\
\hline
\end{tabular}
\end{center}
\label{table:tradeoff}
\end{table}

\section{Conclusions and Future Work}
\label{sec:conclusions}

In this work, we have evaluated the trade-off between accuracy and computational cost for LOMO and XQDA as a classical approach, also for features extracted from the ResNet-50 and MobileNets networks, as a deep learning based method. This evaluation was performed on large-scale datasets, aiming to simulate the scenario of a real-world application. 
In such scenario, the kind of images on which the re-identification is performed, frequently show crowded scenes, which justifies the necessity of having an efficient system that is able to identify as many individuals as possible in the shortest time.

We showed that using features from CNN outperforms by a large margin the accuracy achieved with a classical approach and it is also much faster, when using a GPU. However, this requirement as well as the large amount of annotated data that a network needs to be trained are the drawbacks to consider. Both ResNet-50 and MobileNets achieve a good performance being the second one 4 times faster at test time. Additionally, we proposed network distillation for improving the performance of MobileNets at test time, demonstrating its effectiveness. The student MobileNets networks even outperformed the teacher ResNet-50 model, achieving an accuracy that could not be achieved by training the student independently. 

There are still research lines to explore for the the deep learning case applied to a real scenario. The problem of domain adaptation is still open. It refers to the situation when networks trained with labeled datasets can still perform well with new data recorded in different conditions. 
Also, the retrieval module in the person re-identification pipeline is a bottleneck since a brute-force search is needed in order to compare the person of interest against all the gallery. To solve this, some clustering and indexing approaches have been proposed to reduce the computational cost at test time too, but there is still room for improvement.

\section*{Acknowledgements}
This work was supported by the SURVANT project which has received funding from the European Union’s Horizon 2020 research and innovation programme under grant agreement No 720417. Bogdan Raducanu is supported by Grant No. TIN2016-79717-R, funded by MINECO, Spain.

\section*{References}

\bibliography{mybibfile}

\end{document}